\pdfoutput=1

\documentclass[11pt]{article}

\usepackage[]{naacl2021}

\usepackage{times}
\usepackage{latexsym}
\usepackage{graphicx}

\usepackage[T1]{fontenc}

\usepackage[utf8]{inputenc}

\usepackage{microtype}

\title{TorontoCL at CMCL 2021 Shared Task: RoBERTa with Multi-Stage Fine-Tuning for Eye-Tracking Prediction}

\author{Bai Li$^{1,2}$, Frank Rudzicz$^{1,2,3}$ \\
$^1$ University of Toronto, Department of Computer Science \\
$^2$ Vector Institute for Artificial Intelligence \\ 
$^3$ Unity Health Toronto \\
\texttt{\{bai, frank\}@cs.toronto.edu}
}

\begin{document}
\maketitle
\begin{abstract}
Eye movement data during reading is a useful source of information for understanding language comprehension processes. In this paper, we describe our submission to the CMCL 2021 shared task on predicting human reading patterns. Our model uses RoBERTa with a regression layer to predict 5 eye-tracking features. We train the model in two stages: we first fine-tune on the Provo corpus (another eye-tracking dataset), then fine-tune on the task data. We compare different Transformer models and apply ensembling methods to improve the performance. Our final submission achieves a MAE score of 3.929, ranking 3rd place out of 13 teams that participated in this shared task.
\end{abstract}

\section{Introduction}

Eye-tracking data provides precise records of where humans look during reading, with millisecond-level accuracy. This type of data has recently been leveraged for uses in natural language processing: it can improve performance on a variety of downstream tasks, such as part-of-speech tagging \citep{eyetrack-pos}, dependency parsing \citep{eyetrack-depparse}, and for cognitively-inspired evaluation methods for word embeddings \citep{eyetrack-vec-eval}. Meanwhile, Transformer-based language models such as BERT \citep{bert} and RoBERTa \citep{roberta} have been applied to achieve state-of-the-art performance on many natural language tasks. The CMCL 2021 shared task aims to add to our understanding of how language models can relate to eye movement features.

\begin{figure}
    \centering
    \includegraphics[width=\linewidth]{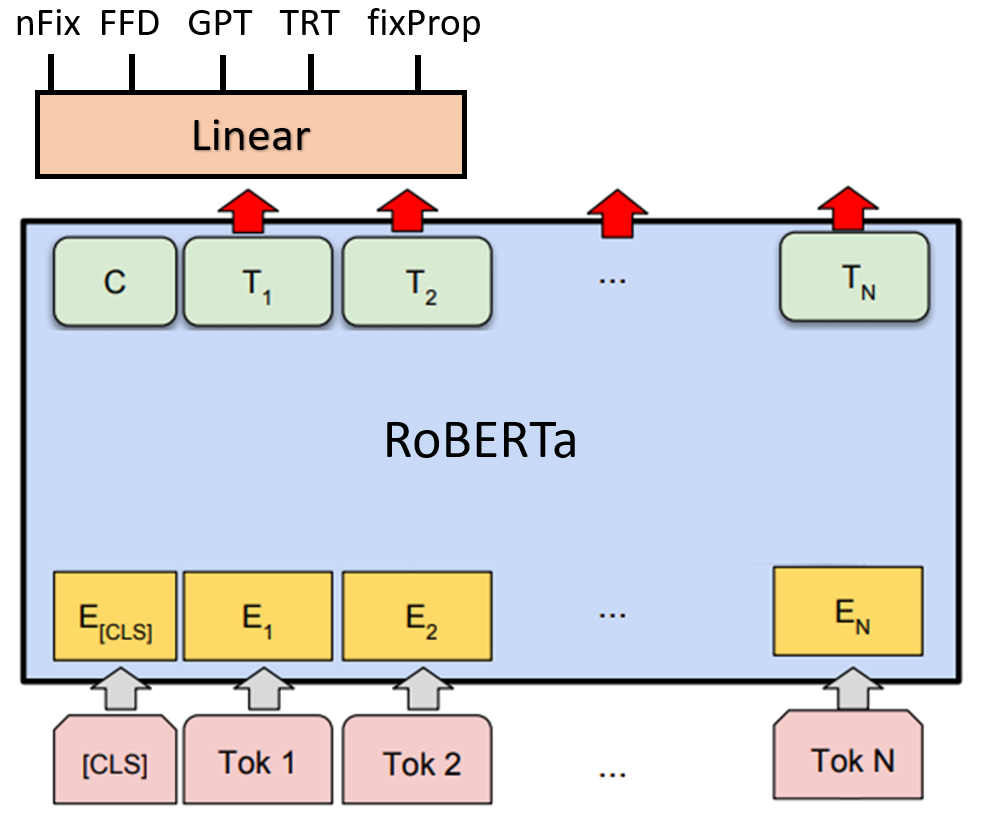}
    \caption{Our model consists of RoBERTa with a regression head on each token, which is a linear layer that predicts the 5 output features from the last layer's embeddings. The model is initialized from pretrained weights and fine-tuned on the task data.}
    \label{fig:roberta-linear}
\end{figure}

In this paper, we present our submission to this shared task, which achieves third place on the leaderboard. We first explore some simple baselines using token-level features, and find that these are already somewhat competitive with the final model's performance. Next, we describe our model architecture, which is based on RoBERTa (Figure \ref{fig:roberta-linear}). We find that model ensembling offers a substantial performance gain over a single model. Finally, we augment the provided training data with the publicly available Provo eye-tracking corpus and combine them using a two-stage fine-tuning procedure; this results in a moderate performance gain. Our source code is available at \url{https://github.com/SPOClab-ca/cmcl-shared-task}.

\section{Task description}

The shared task format is described in \citet{sharedtask}, which we will briefly summarize here. The task data consists of sentences derived from the ZuCo 1.0 and ZuCo 2.0 datasets; 800 sentences (15.7 tokens) were provided as training data and 191 sentences (3.5k tokens) were held out for evaluation. The objective is to predict five eye-tracking features for each token:

\begin{table*}[t]
\centering
\begin{tabular}{lrrrrrr}
\hline
\textbf{Model} & \textbf{nFix} & \textbf{FFD} & \textbf{GPT} & \textbf{TRT} & \textbf{fixProp} & \textbf{All (Dev)} \\ \hline
Median & 7.208 & 1.162 & 3.547 & 2.732 & 21.179 & 7.165 \\
Linear regression     & 4.590 & 0.795 & 2.995 & 1.812 & 13.552 & 4.749 \\
SVR (RBF kernel)      & 4.440 & 0.723 & 2.728 & 1.728 & 12.077 & 4.339 \\ \hline
\end{tabular}
\caption{Baseline results: `Median' is a model that always predicts the median of the training data; the linear regression and SVR models use 4 token-level surface features described in Section \ref{sec:baselines}.}
\label{tab:baselines}
\end{table*}

\begin{table*}[t]
\centering
\begin{tabular}{lrrrrrr}
\hline
\textbf{Model} & \textbf{nFix}  & \textbf{FFD} & \textbf{GPT}   & \textbf{TRT}   & \textbf{fixProp} & \textbf{All (Dev)} \\ \hline
BERT-base     & 4.289 & 0.704          & 2.645 & 1.678 & 11.155 & 4.094 \\
BERT-large    & 4.150 & 0.682          & 2.493 & 1.616 & 11.013 & 3.991 \\
RoBERTa-base   & \textbf{4.066} & \textbf{0.681}        & \textbf{2.443} & \textbf{1.570} & \textbf{10.981}  & \textbf{3.930}   \\
RoBERTa-large & 4.156 & 0.681 & 2.468 & 1.623 & 11.047 & 3.995 \\ \hline
\end{tabular}
\caption{MAE using BERT and RoBERTa models with fine-tuning.}
\label{tab:transformers}
\end{table*}

\begin{table*}[t]
\centering
\begin{tabular}{lrrrrrr}
\hline
\textbf{Model} & \textbf{nFix}  & \textbf{FFD}   & \textbf{GPT}   & \textbf{TRT} & \textbf{fixProp} & \textbf{All (Dev)} \\ \hline
Single Model  & 4.066 & 0.681 & 2.443 & 1.570          & 10.891 & 3.930          \\
Ensemble of 2 & 3.978 & 0.671 & 2.350 & 1.534          & 10.714 & 3.849          \\
Ensemble of 5 & 3.944 & 0.669 & 2.321 & \textbf{1.521} & 10.665 & 3.824 \\
Ensemble of 10 & \textbf{3.943} & \textbf{0.666} & \textbf{2.316} & 1.522        & \textbf{10.660}  & \textbf{3.821}   \\ \hline
\end{tabular}
\caption{Ensembles of RoBERTa-base model, obtained by taking a simple mean of the predictions of individual models. This improves our overall performance by about 0.09 MAE compared to a single model, but with diminishing returns past 5 models.}
\label{tab:ensembles}
\end{table*}

\begin{itemize}
    \setlength\itemsep{-0.25em}
    \item Number of fixations on the current word ({\em nFix}).
    \item First fixation duration of the word ({\em FFD}).
    \item Go-past time: the time from the first fixation of a word until the first fixation beyond it ({\em GPT}).
    \item Total reading time of all fixations of the word, including regressions ({\em TRT}).
    \item Proportion of participants that fixated on the word ({\em fixProp}).
\end{itemize}

The features are averages across multiple participants, and each feature is scaled to be in the range [0, 100]. The evaluation metric is the mean absolute error (MAE) between the predicted and ground truth values, with all features weighted equally.

Since each team is allowed only a small number of submissions, we define our own train and test split to compare our models' performance during development. We use the first 600 sentences as training data and the last 200 sentences for evaluation during development. Except for the submission results (Table \ref{tab:adaptation}), all experimental results reported in this paper are on this development train/test split.

\section{Our approach}

\subsection{Baselines}
\label{sec:baselines}

We start by implementing some simple baselines using token-level surface features. Previous research in eye tracking found that longer words and low-frequency words have higher probabilities of being fixated upon \citep{rayner}. We extract the following features for each token:

\begin{itemize}
    \setlength\itemsep{-0.25em}
    \item Length of token in characters.
    \item Log of the frequency of token in English text, retrieved using the \texttt{wordfreq}\footnote{\url{https://github.com/LuminosoInsight/wordfreq/}} library.
    \item Boolean of whether token contains any uppercase characters.
    \item Boolean of whether token contains any punctuation.
\end{itemize}

Using these features, we train linear regression and support vector regression models separately for each of the 5 output features (Table \ref{tab:baselines}). Despite the simplicity of these features, which do not use any contextual information, they already perform much better than the median baseline. This indicates that much of the variance in all 5 eye-tracking features are explained by surface-level cues.

\begin{table*}[t]
\centering
\begin{tabular}{lrrrrrrr}
\hline
\textbf{Training Data} &
  \textbf{nFix} &
  \textbf{FFD} &
  \textbf{GPT} &
  \textbf{TRT} &
  \textbf{fixProp} &
  \textbf{All (Dev)} &
  \multicolumn{1}{l}{\textbf{Submission}} \\ \hline
Task Only (Single)         & 4.066 & 0.681 & 2.443 & 1.570 & 10.891 & 3.930 & n/a   \\
Provo + Task (Single) & 3.984 & 0.713 & 2.424 & 1.556 & 10.781 & 3.892 & n/a   \\
Task Only (Ensemble)       & 3.943 & 0.666 & 2.316 & 1.522 & 10.660 & 3.821 & 3.974 \\
Provo + Task (ensemble) &
  \textbf{3.888} &
  \textbf{0.664} &
  \textbf{2.306} &
  \textbf{1.499} &
  \textbf{10.586} &
  \textbf{3.789} &
  \textbf{3.929} \\ \hline
\end{tabular}
\caption{Comparison of model trained using the provided versus two-stage fine-tuning using Provo data. The additional pretraining improved overall performance by about 0.04 MAE. Our best submission is an ensemble of 10 RoBERTa-base models with two-stage fine-tuning.}
\label{tab:adaptation}
\end{table*}

\subsection{Fine-tuning transformers}

\begin{figure*}
    \centering
    \includegraphics[width=\linewidth]{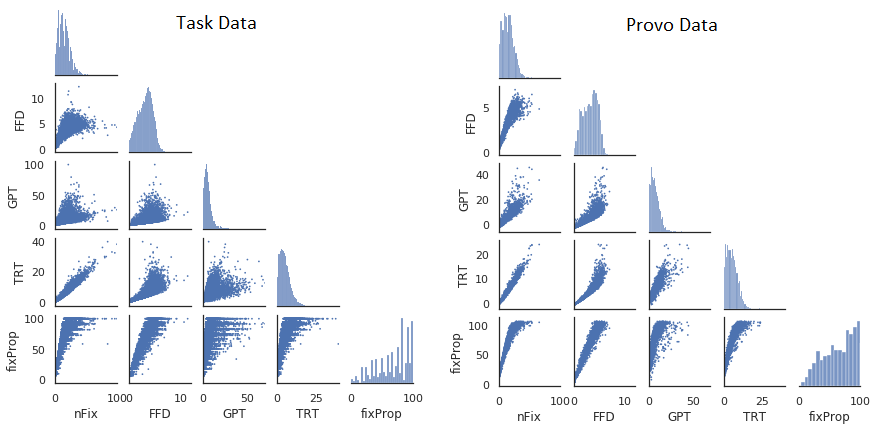}
    \caption{Distributions and pairwise scatterplots of the task data (left) and Provo data processed to match the mean and standard deviation of the task data (right).}
    \label{fig:provo-plots}
\end{figure*}

Our main model uses RoBERTa \citep{roberta} with a linear feedforward layer to predict the 5 output features simultaneously from the last hidden layers of each token. In cases where the original token is split into multiple RoBERTa tokens, we use the first RoBERTa token to make the prediction. The model is initialized with pretrained weights and fine-tuned on the task data to minimize the sum of mean squared errors across all 5 features.

As the task data is relatively small, we found that the model needs to be fine-tuned for 100-150 epochs to reach optimal performance, far greater than the recommended 2-4 epochs \citep{bert}. We trained the model using the AdamW optimizer \citep{adamw} with learning rates of \{1e-5, 2e-5, 5e-5, 1e-4\} and batch sizes of \{8, 16, 32\}; all other hyperparameters were left at their default settings using the HuggingFace library \citep{huggingface}.

In addition to RoBERTa, we experiment with BERT \citep{bert}; we try both the base and large versions of BERT and RoBERTa, using a similar range of hyperparameters for each (Table \ref{tab:transformers}). RoBERTa-base performed the best in our validation experiments; surprisingly, RoBERTa-large had worse performance.

\subsection{Model ensembling}

We use a simple approach to ensembling: we train multiple versions of an identical model using different random seeds and make predictions on the test data. These predictions are the averaged to obtain the final submission. In our experiments (Table \ref{tab:ensembles}), ensembling greatly improves our performance, but with diminishing returns: the MAE of the 10-model ensemble is only marginally better than the 5-model ensemble. We use ensembles of 10 models in our final submission.

\subsection{Domain adaptation from Provo}

In addition to the task data provided, we also use data from the Provo corpus \citep{provo}. This corpus contains eye-tracking data from 84 participants reading 2.6k words from a variety of text sources. The corpus also provides predictability norms and extracted syntactic and semantic features for each word, which we do not use.

We process the Provo data to be similar to the task data so that they can be combined. First, we identify the Provo features that are most similar to each of the output features: we map {\em IA\_FIXATION\_COUNT} to {\em nFix}, {\em IA\_FIRST\_FIXATION\_DURATION} to {\em FFD}, {\em IA\_REGRESSION\_PATH\_DURATION} to {\em GPT}, and {\em IA\_DWELL\_TIME} to {\em TRT}, taking the mean across all participants for each feature. For the {\em fixProp} feature, we calculate the proportion of participants where $IA\_DWELL\_TIME > 0$ for each word. Finally, we scale all five features to have the same mean and standard deviation as the task data, and verify that their distributions and pairwise scatterplots are similar (Figure \ref{fig:provo-plots}).

We use two-stage fine-tuning to combine the Provo data with the task data. In two-stage fine-tuning, the entire model is fine-tuned on an auxiliary task before fine-tuning on the target task -- this often yields a performance improvement, especially when the target task has a small amount of data \citep{intermediate-fine-tuning}. In our case, we fine-tune the RoBERTa-base model for 100 epochs on the Provo data, then fine-tune for another 150 epochs on the task data. This gave a considerable improvement on both the development and submission scores (Table \ref{tab:adaptation}). Our best final submission is an ensemble of 10 identical models trained this way with different random seeds.

\section{Conclusion and future work}

We propose a simple approach to predict eye-tracking features using the RoBERTa model customized with a per-token regression head. Our initial model uses the standard fine-tuning procedure; experiments show that the performance is further improved by model ensembling and domain adaptation by two-stage fine-tuning on an intermediate eye-tracking task. Our best model achieves third place on the leaderboard.

In future work, several avenues may be explored to further improve performance. First, we did not combine our feature engineering baseline with the RoBERTa model -- engineered features (such as frequency statistics or neurolinguistic norms) would provide the model with information not contained in RoBERTa. Second, we only experimented with a small subset of features from the Provo corpus for domain adaptation, whereas it is not actually necessary for the auxiliary fine-tuning task to match the target task. Thus, it may be possible to achieve better performance by fine-tuning on a different set of Provo features, or a different dataset entirely.

\section*{Acknowledgements}

We thank Dr Jeanne Sinclair for her helpful discussions during this project. FR is supported by a CIFAR Chair in Artificial Intelligence.

\bibliography{custom}

\begin{thebibliography}{11}
\expandafter\ifx\csname natexlab\endcsname\relax\def\natexlab#1{#1}\fi

\bibitem[{Barrett et~al.(2016)Barrett, Bingel, Keller, and
  S{\o}gaard}]{eyetrack-pos}
Maria Barrett, Joachim Bingel, Frank Keller, and Anders S{\o}gaard. 2016.
\newblock Weakly supervised part-of-speech tagging using eye-tracking data.
\newblock In \emph{Proceedings of the 54th Annual Meeting of the Association
  for Computational Linguistics (Volume 2: Short Papers)}, pages 579--584.

\bibitem[{Devlin et~al.(2019)Devlin, Chang, Lee, and Toutanova}]{bert}
Jacob Devlin, Ming-Wei Chang, Kenton Lee, and Kristina Toutanova. 2019.
\newblock {BERT}: Pre-training of deep bidirectional transformers for language
  understanding.
\newblock In \emph{Proceedings of the 2019 Conference of the North American
  Chapter of the Association for Computational Linguistics: Human Language
  Technologies, Volume 1 (Long and Short Papers)}, pages 4171--4186.

\bibitem[{Hollenstein et~al.(2021)Hollenstein, Chersoni, Jacobs, Oseki,
  Prévot, and Santus}]{sharedtask}
Nora Hollenstein, Emmanuele Chersoni, Cassandra Jacobs, Yohei Oseki, Laurent
  Prévot, and Enrico Santus. 2021.
\newblock {CMCL} 2021 shared task on eye-tracking prediction.
\newblock In \emph{Proceedings of the Workshop on Cognitive Modeling and
  Computational Linguistics}.

\bibitem[{Liu et~al.(2019)Liu, Ott, Goyal, Du, Joshi, Chen, Levy, Lewis,
  Zettlemoyer, and Stoyanov}]{roberta}
Yinhan Liu, Myle Ott, Naman Goyal, Jingfei Du, Mandar Joshi, Danqi Chen, Omer
  Levy, Mike Lewis, Luke Zettlemoyer, and Veselin Stoyanov. 2019.
\newblock {RoBERTa}: A robustly optimized {BERT} pretraining approach.
\newblock \emph{arXiv preprint arXiv:1907.11692}.

\bibitem[{Loshchilov and Hutter(2018)}]{adamw}
Ilya Loshchilov and Frank Hutter. 2018.
\newblock Decoupled weight decay regularization.
\newblock In \emph{International Conference on Learning Representations}.

\bibitem[{Luke and Christianson(2018)}]{provo}
Steven~G Luke and Kiel Christianson. 2018.
\newblock The {P}rovo corpus: A large eye-tracking corpus with predictability
  norms.
\newblock \emph{Behavior research methods}, 50(2):826--833.

\bibitem[{Pruksachatkun et~al.(2020)Pruksachatkun, Phang, Liu, Htut, Zhang,
  Pang, Vania, Kann, and Bowman}]{intermediate-fine-tuning}
Yada Pruksachatkun, Jason Phang, Haokun Liu, Phu~Mon Htut, Xiaoyi Zhang,
  Richard~Yuanzhe Pang, Clara Vania, Katharina Kann, and Samuel Bowman. 2020.
\newblock Intermediate-task transfer learning with pretrained language models:
  When and why does it work?
\newblock In \emph{Proceedings of the 58th Annual Meeting of the Association
  for Computational Linguistics}, pages 5231--5247.

\bibitem[{Rayner(1998)}]{rayner}
Keith Rayner. 1998.
\newblock Eye movements in reading and information processing: 20 years of
  research.
\newblock \emph{Psychological bulletin}, 124(3):372.

\bibitem[{S{\o}gaard(2016)}]{eyetrack-vec-eval}
Anders S{\o}gaard. 2016.
\newblock Evaluating word embeddings with {fMRI} and eye-tracking.
\newblock In \emph{Proceedings of the 1st Workshop on Evaluating Vector-Space
  Representations for NLP}, pages 116--121.

\bibitem[{Strzyz et~al.(2019)Strzyz, Vilares, and
  G{\'o}mez-Rodr{\'\i}guez}]{eyetrack-depparse}
Michalina Strzyz, David Vilares, and Carlos G{\'o}mez-Rodr{\'\i}guez. 2019.
\newblock Towards making a dependency parser see.
\newblock In \emph{Proceedings of the 2019 Conference on Empirical Methods in
  Natural Language Processing and the 9th International Joint Conference on
  Natural Language Processing (EMNLP-IJCNLP)}, pages 1500--1506.

\bibitem[{Wolf et~al.(2020)Wolf, Debut, Sanh, Chaumond, Delangue, Moi, Cistac,
  Rault, Louf, Funtowicz et~al.}]{huggingface}
Thomas Wolf, Lysandre Debut, Victor Sanh, Julien Chaumond, Clement Delangue,
  Anthony Moi, Pierric Cistac, Tim Rault, R{\'e}mi Louf, Morgan Funtowicz,
  et~al. 2020.
\newblock Transformers: State-of-the-art natural language processing.
\newblock In \emph{Proceedings of the 2020 Conference on Empirical Methods in
  Natural Language Processing: System Demonstrations}, pages 38--45, Online.
  Association for Computational Linguistics.

\end{thebibliography}
\bibliographystyle{acl_natbib}

\end{document}